\documentclass{article}

     \usepackage[final]{neurips_2019}

\usepackage[utf8]{inputenc} % allow utf-8 input
\usepackage[T1]{fontenc}    % use 8-bit T1 fonts
\usepackage{hyperref}       % hyperlinks
\usepackage{url}            % simple URL typesetting
\usepackage{booktabs}       % professional-quality tables
\usepackage{amsfonts}       % blackboard math symbols
\usepackage{nicefrac}       % compact symbols for 1/2, etc.
\usepackage{microtype}      % microtypography
\usepackage{times}
\usepackage{latexsym}
\usepackage{graphicx}
\usepackage{amsfonts}
\usepackage{amsmath}
\usepackage{tikz}
\usepackage{pgfplots}
\usepackage{verbatim}
\usepackage{multirow}
\usepackage{booktabs}
\usepackage{subfigure}
\usepackage{algorithm}
\usepackage{algorithmicx}
\usepackage{algpseudocode}

\title{Knowledge Graph Completion with Text-aided Regularization}

\author{%
  Tong Chen \ Sirou Zhu \ Yiming Wen \ Zhaomin Zheng \\
  Language Technology Institute\\
  Carnegie Mellon University\\
  Pittsburgh, PA 15213 \\
  \texttt{\{tongc2, sirouz, yimingwe, zhaominz\}@cs.cmu.edu} \\
}

\begin{document}

\maketitle

\begin{abstract}
  Knowledge Graph Completion is a task of expanding the knowledge graph/base through estimating possible entities, or proper nouns, that can be connected using a set of predefined relations, or verb/predicates describing interconnections of two things. Generally, we describe this problem as adding new edges to a current network of vertices and edges. Traditional approaches mainly focus on using the existing graphical information that is intrinsic of the graph and train the corresponding embeddings to describe the information; however, we think that the corpus that are related to the entities should also contain information that can positively influence the embeddings to better make predictions. In our project, we try numerous ways of using extracted or raw textual information to help existing KG embedding frameworks reach better prediction results, in the means of adding a similarity function to the regularization part in the loss function. Results have shown that we have made decent improvements over baseline KG embedding methods.
\end{abstract}

\section{Introduction}
The Knowledge Graph/Knowledge Base is a collection of structured tuples of information connecting entities via relations, where \textbf{entities} are generalized concepts or names to be appeared in head/tail positions, forming nodes/vertices in the graph and \textbf{relations} are verbs or predicates describing the way two entities are logically connected, forming edges in the graph.

Formally, Knowledge Graph can be represented as a directed multigraph that may have multiple edges with the same vertices. 

Knowledge Graphs have a lot of applications, such as Structured Search, Question-Answering Systems, Recommendation Systems etc. However, most of the Knowledge Graphs are large but far from complete. In this project, we aim to provide a modern approach to create or predict missing links in the graph, formally done by link prediction: given two elements of a triple, predict missing items. 

Let's formally define our task. Let $\mathcal{E}$ be be the set of entities and $\mathcal{R}$ be the set of relations. Then a KG is defined as a collection of triple facts ($\mathit{e_s,r,e_d}$), and the task of knowledge graph reasoning is to find the missing entry in the triple facts. 

Previously, the focus of knwoledge graph embedding and completion mainly focuses on using various methods to catch the intrinsic, latent semantics of the knowledge graph, which more or less rely on the existing items or structures of the graph itself; but it is another level of concern that the information would be limited to the existing relations or semantics of the knowledge graph, so to complete and expand a knowledge graph, it would be very helpful to introduce external information from text corpora that one knowledge graph or even commonsense graph would rely on, be it general (like FreeBase or YAGO) or industry-specific (such asUMLS).  

Many recent joint graph and text embedding methods have been focusing on learning better knowledge graph embeddings for reasoning (\cite{ OpenNRE:18}), but we consider reaching for better graph embeddings in a more language-oriented sense. In this research, we would propose a general framework of using regularization techniques to train a set of entity embeddings that can capture the nuances of relation and connection between the entities that may not lie in the knowledge graph structure itself, but more on the text corpus side; through becoming a constituent part of regularization, we make our framework more compatible to more of the commonly available knowledge graph embedding models, and have the potential to migrate to state-of-the-art models of the time.

\section{Related Work}

\subsection{Knowledge Graph Reasoning}
Given a knowledge graph, the task of knowledge graph reasoning includes predicting missing links between entities, predicting missing entities, and predicting whether a graph triple is true or false. A variety of methods have been applied to the task of knowledge graph reasoning, and recently embedding-based methods are gaining popularity and yielding promising results, such as linear models in \cite{TransE:11}, matrix factorization models in \cite{ComplEx:16}, convolutional neural networks in \cite{DistMult:15}. In spite of promising results, these models have limited interpretability. Reinforcement learning models have better interpretability, such MINERVA in \cite{das2017go}, DeepPath in \cite{xiong2017deeppath} and Multi-hop in \cite{multi-lin2018}, which exploit policy network. 

\subsection{Open-world Knowledge Graph Completion.}
Using only information inside the static and structured knowledge graph limits our ability to learn representations for embeddings and relations. Therefore, (\cite{shi2018open}) raised the concept of Open-world Knowledge Graph Completion problem, which is to utilize external information so as to connect unseen entities to the knowledge graph. There are several attempts to utilize text corpus to improve Knowledge Graph embeddings, such as the mutual attention model by (\cite{OpenNRE:18}) and Latent Relation Language Models in \cite{hayashi2019latent}. However, these approaches have the following limitations: First of all, the models focuses on local information in the text corpus. Each entity is trained with just its description without considering corpus-level statistics. Secondly, the model need to align two disjoint latent spaces, the Knowledge Graph space and the text corpus space. Thirdly, the models tend to closely couple between the Knowledge Graph and Text method. Last but not least, the models typically provide no way of learning relation-specific representations of entities.

\subsection{Text Embedding Models and Vector Sets.} 
Word embeddings have been proved to be useful in NLP tasks as standalone features (\cite{turian-etal-2010-word}). The key idea of word embedding is to use multi-dimentional vectors to represent the meanings of words. Influential models can be divided into two categories, count-based models and prediction-based models. A good example of count-based models is GloVe, introduced by Pennington
et al. in 2014, which is a log-linear model trained on window based local co-occurrence
information about word pairs in \cite{jeffreypennington2014glove}. Popular prediction-based models include Continuous bag
of words (CBOW) models, skip-gram models and transformer-based models. Based on these state-of-the-art models, pre-trained vector sets such as Word2Vec in  \cite{mikolov2013distributed}, FastText in  \cite{mikolov2018advances},
and GloVe, are released and are ready to be used in NLP tasks.

In our model, word embedding can be used both statically and dynamically. The static way is to initialize the entity weight by pretrained embedding from results in \cite{jeffreypennington2014glove}, \cite{mikolov2013distributed}, and train the entity embedding with models in \cite{conve:18} and \cite{TransE:11}. The dynamic way is to combine the loss functions of word embedding model and knowledge graph model together to train the entity and relation embedding.

\section{Text-regularized KG Completion}

 In this research, we would propose a general framework of using regularization techniques to train a set of entity embeddings that can capture the nuances of relation and connection between the entities that may not lie in the knowledge graph structure itself, but more on the text corpus side. Many recent joint graph and text embedding methods have been focusing on learning better knowledge graph embeddings for reasoning in \cite{OpenNRE:18}, but we consider reaching for better graph embeddings in a more language-oriented sense.

Existing jointly training methods, like \cite{Danqi:15}, \cite{OpenNRE:18}, and \cite{hayashi2019latent}, share the following shortcomings: 

\begin{itemize} 

\item These methods typically only focus on a neighboring/localized space that is labeled relevant to some given entity-relation-entity triples; their embeddings are only trained with the related sentence description that are labeled in the corpus.

\item They do not take corpus level statistics and language model relativities into account, due to the same reason that they rely on labeled partitioned datasets. 

\item They need to align two disjoint latent/embedding spaces (the KG embeddings and text embeddings), which are usually not quite overlapping considering the word tokens available from both sides.

\item Entities representations are typically generic, not accounting for relation variations, so no way of learning relation-specific embeddings are offered.
\end{itemize}

Here we would propose a method of text-enhanced knowledge graph embedding, which uses similarity functions as regularizers towards the training loss of the knowledge graph. The underlied motivation is that, we would typically find similar descriptions and especially, overlapping words for two entities/concepts that falls into similar domains and same categories of name types; in the meantime, these entities would also share similar characteristics in their connectivity, topology, and types of relations linking to them in the knowledge graph. Thus from the description or context regarding the concept, we would be able to train a similar set of embeddings for these two concepts. 

For example, both \textbf{Microsoft} and \textbf{Amazon} are concept names that describes a company, and their descriptions would definitely cover the fact that they both \textit{have headquarters} in \textbf{Seattle}, and they both \textit{have a founder} (who are \textbf{Bill Gates} and \textbf{Jeff Bezos}, respectively).   

\begin{figure}[tb]
\centering
\vspace{-0.1cm}
\includegraphics[width=0.6\columnwidth]{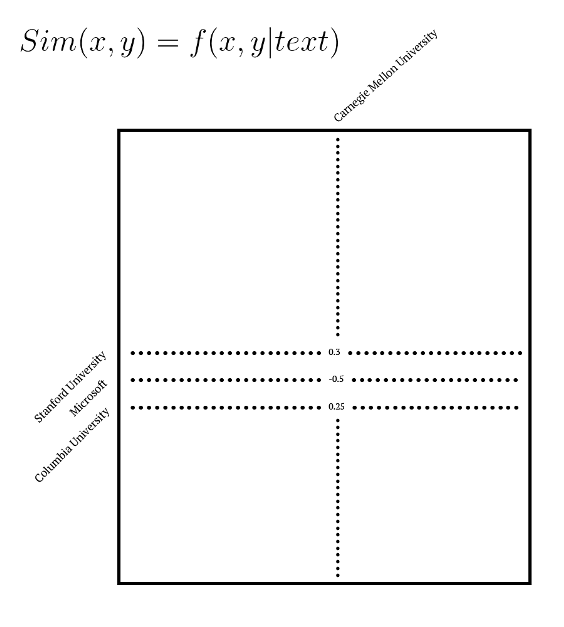}
\vspace{-0.2cm}
\caption{\textbf{Illustration of a similarity function over the corpus space.} Given two entities (e.g., \textbf{Carnegie Mellon University} and \textbf{Stanford}, or \textbf{Pittsburgh}), we calculate the similarities between them using any defined function that is defined across the text corpus.}
\label{fig:Example}
\vspace{-0.2cm}
\end{figure}

To learn a corpus-regularized representations for the relations and entities, we would need to build a loss function that properly trains the embedding towards better predictability. we could define a loss function $L_(text) = Sim(x, y) = f(x, y | text)$ on the domain $x, y \in \mathbf{R} ^{|E| * |E|}$ that captures the similarities between relevant descriptions or context of entities $x$ and $y$, as shown in the figure \ref{fig:Example}. There are lots of room of decision over which type of similarity function $f$ to be used, such as word overlap on context, TF-IDF based on word pairs, dot-product of existing trained embeddings, and so on.

Formally, we would define a new loss function that serves the purpose of minimizing losses from both the side of knowledge base and the one from the text:

\begin{center}
$\mathit{L} = \sum_{(\mathbf{e_1, e_2, r}) \in KG} \lambda_1 \times \mathit{L}_{KG} (\mathbf{e_1, e_2, r}) + \lambda_2 \times \mathit{L}_{Text}(\mathbf{e_1, e_2})$
\end{center}

To exemplify one possible similarity function, we could propose a more GLoVe\cite{jeffreypennington2014glove}-like approach, which could make entities that have similar descriptions being closer to each other in the vector embedding space. 

\begin{center}
$\mathit{L} = \sum_{(\mathbf{e_1, e_2, r}) \in KG} \lambda_1 \times \mathit{L}_{KG} (\mathbf{e_1, e_2, r}) + \lambda_2 \times (\mathbf{e_1^Te_2 + b_{e1} + b_{e2} - X_{e_1, e_2}})$
\end{center}

\section{Variants of Similarity functions}
Here we show some types of similarity functions that we have tested with. They include converging to generic word embeddings, similarity calculation using entity-recognized newspaper text, and ones using associated wikipedia article text. 

\subsection{Variant A: Converging to existing embeddings}
A Quick way to use existing fruits of extracted textual information to begin with is converging to existing word embeddings. Given the premises that we are trying to incorporate textual structures into entity embeddings, we can well assume that word embeddings should also contain information about connections between words. 

Considering the apparent fact that the latent structures underlying in the knowledge graphs should be drastically different than the ones extracted from text (which could be more focusing on predicting the next words that would follow the current word), we would prefer to converge the differences between the embeddings of the entities, and the ones when they are converted to words. Also considering the fact that the latent structures' distribution and the number of dimensions should be different between the two types of embeddings, we incorporate a fully connection layer to extract and redistribute the features, and make the differences comparable.

The easiest and the most prominent example would be using cosine similarity to converge the two:

\begin{align*}
    \mathbf{Sim}(\mathbf{e_1}, \mathbf{e_2}) = \cos(E_\mathbf{e_1} - E_\mathbf{e_2},f(w_\mathbf{e_1} - w_\mathbf{e_2})) \\
    f(w_\mathbf{e_1} - w_\mathbf{e_2}) = NN(w_\mathbf{e_1} - w_\mathbf{e_2})
\end{align*}

Another would be using rank distance defined by (\cite{santus2018rank}) that would extract the most prominent ranks of the embeddings and compare the differences between the two:

\begin{align*}
    \mathbf{Sim}(\mathbf{e_1}, \mathbf{e_2}) = \sum_{r \in \text{intersect}} \frac{2}{(E_\mathbf{e_1} - E_\mathbf{e_2} + f(w_\mathbf{e_1} - w_\mathbf{e_2})} \\
    f(w_\mathbf{e_1} - w_\mathbf{e_2}) = NN(w_\mathbf{e_1} - w_\mathbf{e_2})
\end{align*}

\subsection{Variant B: Using Entity-tagged Text}
In (\cite{jeffreypennington2014glove}), word similarities can be extracted from a corpus by constructing a co-occurrence matrix. Similarly, entity similarities can be calculated in the same manner. Given a corpus, we can construct matrix $\mathbf{X}$ where $\mathbf{X_{i,j}}$ is the number of times entity $\mathbf{j}$  occurs in the context of entity $\mathbf{i}$. We select New York Time corpus as the test dataset for extracting co-occurrence matrix $\mathbf{X}$. Summing over the entities in the knowledge graph, we have our Entity-tagged text regularization term:
\begin{align*}
    \mathit{L}_{Entity-tagged} \sum_{(\mathbf{e_1, e_2, r}) \in KG} \mathbf{w_i^T w_j - log X_{i,j}}
\end{align*}

\subsection{Variant C: Using Associated Wikipedia Text}
As described above, similar entities should have similar text descriptions on Wikipedia and similar text structures. For example, Wikipedia documents of counties in the US should all talk about its climate, economy, population, etc, and actors should all talk about their career works and life experiences and so on. To exploit such similarity to regularize our entity embedding, we measure two correlations between the Wikipedia documents of entities, which are, first, a modified Relaxed Word Mover Distance (\cite{kusner2015word}) between two entity documents, and second, TF-IDF similarity between two entity documents.

\subsubsection{Relaxed Word Mover Distance}
In (\cite{kusner2015word}), the Relaxed Word Mover Distance is calculated as $$\text{RWMD}(\mathbf{e_1, \mathbf{e_2}}) = \sum_{i \in \mathbf{e_1}} c_i \min_{j \in \mathbf{e_2}}dist(i, j)$$, where  $i, j$ are index of words in entity documents, $c_i$ is the normalized frequency of word $i$, and $dist(i, j)$ is the distance between the pre-trained word embedding of word $i$ and $j$. In our proposed method, to boost the calculation, instead of using the distance between word embedding, we use the the dot product of two word embedding to measure their similarity, and accordingly, we introduce the concept of RWMD-Gain, which is the maximum gain of transformation from the document of an entity to another and is as follows:
\begin{align*}
\text{RWMD-Gain}(\mathbf{e_1, e_2}) = \sum_{i \in \mathbf{e_1}} c_i \max_{j \in \mathbf{e_2}}sim(i, j)
\end{align*}

We employ the bidirectional gain as the semantic similarity of two entity document, which is 
\begin{align*}
    \text{Gain}(e_1, e_2) = \text{RWMD-Gain}(e_1, e_2) + \text{RWMD-Gain}(e_2, e_1)
\end{align*}

To sum over all instances in the knowledge graph, we have our RMWD regularization term:
\begin{align*}
    \mathit{L}_{Text-RWMD} = 
    \sum_{(\mathbf{e_1, e_2, r}) \in KG}log(\text{Gain}(\mathbf{e_1}, \mathbf{e_2}))|\mathbf{e_1}-\mathbf{e_2}|^2_2
\end{align*}

\subsubsection{TF-IDF Similarity}
To compute the TF-IDF similarity between two entity documents, we first convert each document $\mathit{d}$ to a TF-IDF vector $\mathbf{w_d}$, in which each element can be calculated by the following formula:
\begin{align*}
    w_{i,d} = tf_{i,d} \times log(\frac{N}{df_i})
\end{align*}
where $\mathit{i}$ represents the index of the word; $tf_{i,d}$ represents the occurrence of word $\mathit{i}$ in document $\mathit{d}$; $\mathit{N}$ represents the total number of documents and $df_i$ represents the number of documents containing word $\mathit{i}$.

The similarity of document $d_x$ and document $d_y$ is the dot product of such $\mathbf{w_x}$, $\mathbf{w_y}$.
\begin{align*}
    Sim(x, y) = dot(\mathbf{w_x}, \mathbf{w_y})
\end{align*}
% $sim(i, j)=w_iw_j$, where $w_i$ represents the pretrained GloVe word embedding of word $i$.

% Finally, we would incorporate relation-sensitive entity-to-entity similarity functions, so that the similarity between different entities could be calculated accordingly when a given relation is designated. Through this we can link some entities towards some particular set of relations by amplifying the embedding values of some entities in some relations, while being suppressed in others. For example, we could make entities that are location names, for example, \textbf{Pittsburgh} or \textbf{Wuhan}, more possible of being a tail/object entity of the relation \textit{locatedIn}, \textit{bornin} or \textit{headquarteredIn}.

% \begin{center}
% $\mathit{L} = \sum_{(\mathbf{e_1, e_2, r}) \in KG} \lambda_1 \times \mathit{L}_{KG} (\mathbf{e_1, e_2, r}) + \lambda_2 \times \mathit{L}_{Text}(\mathbf{e_1, e_2 | r})$
% \end{center}

\section{Experiments}

\subsection{Data Sources and formulations}

From (\cite{fu2019collaborative}) we select the \textbf{FB60K} dataset as the base knowledge graph \footnote{\small\url{https://github.com/thunlp/OpenNRE}}. In the aforementioned paper, the authors studied the datasets and found that the relation distributions of the two datasets are very imbalanced; but still, It is the KG dataset that contains the most entities in the “FreeBase”-related KG variants, like FB17K, FB15K-237, etc. In the original dataset, all entity names are its Freebase ID; to properly link the partially anonymized graph to the truth raw text, we employed an openly available dictionary to convert thme from ID to its corresponding names. By linking the ID to the names, we can see that the FB60K dataset mainly contains famous or obscure locations, celebrity names, schools, sports unions, etc. 

On the external sources of the three variants, we would list them here:
\begin{itemize} 

\item \textbf{Variant A (using existing embeddings)}: we employ the GLoVe trained embeddings (\cite{pennington2014glove}) as the embeddings we would want to converge to. Specifically, since word embeddings are normally single words, while entities are mostly proper noun phrases, we generate the final embedding of the entity by calculating the average over all the constituent words that is related to the entity (an entity can have multiple words and multiple names to refer to) according to the open dictionaries.

\item \textbf{Variant B (using entity labeled text)}: we employ the NYT10 dataset that is supplied along with the FB60K dataset. Entities are labeled using common named entity recognition toolkits, and the source corpus text is excerpted from the New York Times corpus. They are also labeled what relations the sentence would contain, but currently we haven't employed these. An analysis to the information overlap (i.e., alignment) between the corpus and the KG in Table \ref{tab:dataoverlap}. Higher CT/CE (triple-entity ratio) indicates adding corpus-edges to the KG increases the average degree more significantly, leading to more reduction in sparsity.

\item \textbf{Variant C (using linked Wikipedia articles)}: we employ the articles from the Wikipedia dump in December 2019. By linking the entity FreeBase ID to its name, and by finding these names' English Wikipedia article, we managed to extract over 54,000 articles linking to existing entities (around 80\%). We use Stanford CoreNLP Tokenizer (\cite{manning-EtAl:2014:P14-5}) to tokenize and lower-letter all the words in the text after we remove structured text and tables, which are text that cannot be easily consumed. 
\end{itemize}

We choose TransE (\cite{TransE:11}) as the base KG embedding method as the base of comparison between all the variants. TransE is an old methods of calculating KG embeddings, but it is the easiest to implement and has still been holding decent performances despite several newer and more state-of-the-art models being introduced since. 

  \begin{table*}[tb]
    \centering
    \small
    \scalebox{0.77}{
    \begin{tabular}{ccccccccccc}
         \hline
         Dataset & \#triples(C) & \#triple(G) & \#entities(C) & \#entities(G) & \#rel(C) & \#rel(G) &S(train) &S(test) & CT/CE & CR/KR\\
         \hline
         FB60K-NYT10  & 172,448 & 268,280 &63,696 & 69,514 &57&1,327&570k & 172k & 2.71 & 0.04\\
         \hline
    \end{tabular}}
    \vspace{-0.1cm}
    \caption{\textbf{The dataset information.} \#triples(C) \& \#triples(G) denote the number of triples in the corpus and the KG respectively, and so on. S(train) denotes the number of sentences in the training corpus, while S(test) denotes the number of sentences in the testing corpus. CT/CE denotes triple-entity ratio. Lower triple-entity ratio indicates less triples per entity in average can be extracted from the corpus. CR/KR denotes corpus-relation-quantity/KG-relation-quantity ratio. Lower CR/KR indicates less information overlap between the corpus and the KG.}
    \label{tab:dataoverlap}
\end{table*}

\subsection{Experimental Result}
We run all of our proposed variants on FB60K-NYT10 dataset. Our baseline method is TransE (\cite{TransE:11}). We add different proposed regularization terms to TransE individually to compare their performance. The experimental results are shown in Table \ref{tab:experiment}.

\begin{table}[h]
 \centering
    \small
    \scalebox{0.77}{
\begin{tabular}{lcccccccc}
\hline
{ Method}             & { Hits@1}               & { Hits@3}               & { Hits@10}              & { MRR}               & { Hits@1 Filtered}      & { Hits@3 Filtered}      & { Hits@10 Filtered}     & { MRR Filtered}     \\ \hline
{TransE}            & { 0.2144}          & { 0.4414}          & { 0.5431}          & { 0.3379}          & { 0.3392}          & { 0.6245}          & { 0.7016}          & { 0.4902}         \\ 
{ TransE  + Cosine}     & { 0.2303}          & { 0.4455}          & { 0.5436}           & { 0.3487}          & { 0.3957}          & { \textbf{0.6310}} & { \textbf{0.7035}} & { \textbf{0.5216}} \\ 
{ TransE + GloVe-NYT}  & { 0.2112}          & { 0.4366}          & { 0.5436}          & { 0.3363}           & { 0.3378}            & { 0.6086}          & { 0.6997}          & { 0.4841}         \\ 
{ TransE + GloVe-Wiki} & { 0.2305}          & { \textbf{0.4474}} & { \textbf{0.5493}} & { \textbf{0.3500}} & { 0.3552}          & { 0.6170}          & { 0.6951}          & { 0.4957}         \\ 
{ TransE + GloVe-RWMD} & { \textbf{0.2417}} & { 0.4000}          & { 0.5147}          & { 0.3396}            & { \textbf{0.4292}} & { 0.5816}          & { 0.6787}           & { 0.5194}         \\ \hline
\end{tabular}}
\caption{\textbf{Experimental results} running on FB60K-NYT10 dataset. The best of all metrics are highlighted with bold.}
\label{tab:experiment}
\end{table}

% \section{Objectives}

% \begin{itemize}
%     \item Develop a method of KG-text joint training that does not rely on entity and relation extracted corpora, and extract information from raw texts, evaluating entity similarities on raw text measures.
%     \item Inspect upon various possibilities of evaluating loss functions and similarity functions, making sure to employ state-of-the-art current methods and reach for a better performance compared to pure KG-based embedding methods
%     \item Discuss the possibilities of generating relation-sensitive embeddings for entities, in order to further boost performance of KG completion using our current method
    
% \end{itemize}
\subsection{Experimental Analysis}
From table 2, we can see that generally TransE + Cosine and TransE + GloVe-wiki outperform the other methods and baseline. For HR@1, TranE + GloVe-RWMD gives the best result and outperforms baseline significantly. For HR@3, HR@10, and MRR, TransE + GloVe is the best one. GloVe-RWMD considers the semantic distances, which helps finding the best suitable entity, but negatively impacts the score of the potential entities. TransE + Cosine and GloVe-wiki are more simple and intuitive, which consider only the word frequencies and proves to be more useful when considering a list of entities. The amount of injected information directly influences similarity matrix quality. Wiki is comparably larger then New York Times, and larger amount of information gives more accurate entity similarity score, so GloVe-Wiki outperforms  Glove-NYT.

\section{Conclusion}

From the experiments, our methods showed decent improvements compared to the baseline model. Although TransE is a prudent model, we believe the proposed regularization methods can fit to later state-of-the-art models with modest adaptations. However, due to time limitation, the hyper-parameters were not fine-tuned in the experiments so the results did not show a very significant improvement. Therefore, future work can be done on further improving the implementation of regularization, or on choosing a better set of parameters. Moreover, the design of decoders can be further improved to transform textual latent features to entity or knowledge graph latent features. After all, we believe that more information should always be better than less, but we need to find a good way to utilize it. 

In this project, we diversely explored text-assisted KG embedding or reasoning methods. However, given the fact that these methods all require extensive frameworks to extract external information, we think there might be unexplored while simpler ways to migrate textual latent features to enrich KG embeddings. However, further study requires us to focus on a more balanced and diverse knowledge base, which we are lacking today due to the demise of the FreeBase graph.

\bibliography{ref}
\bibliographystyle{ref}

\end{document}